\documentclass[runningheads]{llncs}
\usepackage[T1]{fontenc}
\usepackage{graphicx}
\usepackage{siunitx,amssymb,amsmath,booktabs,multirow,soul,hyperref}
\DeclareMathOperator{\diag}{diag}

\newcommand{\revision}[2]{#2}

\begin{document}
\title{GReAT: leveraging geometric artery data to improve wall shear stress assessment}
\titlerunning{Geometric representation adaption transformer}
\author{Julian Suk\inst{1,2} \and
Jolanda J. Wentzel\inst{3} \and
Patryk Rygiel\inst{4} \and
Joost Daemen\inst{3} \and
Daniel Rueckert\inst{1,2,5} \and
Jelmer M. Wolterink\inst{4}}
%\author{First Author\inst{1}\orcidID{0000-1111-2222-3333} \and
%Second Author\inst{2,3}\orcidID{1111-2222-3333-4444} \and
%Third Author\inst{3}\orcidID{2222--3333-4444-5555}}
%
\authorrunning{J. Suk et al.}
\institute{Technical University of Munich, Munich, Germany\\
\email{julian.suk@tum.de} \and
Munich Center for Machine Learning, Munich, Germany \and
Erasmus Medical Center, Rotterdam, The Netherlands \and
University of Twente, Enschede, The Netherlands \and
Imperial College London, London, United Kingdom}
%\institute{Princeton University, Princeton NJ 08544, USA \and
%Springer Heidelberg, Tiergartenstr. 17, 69121 Heidelberg, Germany
\maketitle              % typeset the header of the contribution
\begin{abstract}
Leveraging big data for patient care is promising in many medical fields such as cardiovascular health.
For example, hemodynamic biomarkers like wall shear stress could be assessed from patient-specific medical images via machine learning algorithms, bypassing the need for time-intensive computational fluid simulation.
However, it is extremely challenging
% for hospitals
to amass large-enough datasets to effectively train such models.
We could address this data scarcity by means of
%transfer learning,
self-supervised pre-training
%, representation learning
and foundations models given large datasets of geometric artery models.
In the context of coronary arteries, leveraging learned representations to improve hemodynamic biomarker assessment has not yet been well studied.
In this work, we address this gap
%in the literature
by investigating whether a large dataset (8449 shapes) consisting of geometric models of 3D blood vessels can benefit wall shear stress assessment in coronary artery models from a small-scale clinical trial (49 patients).
We create a self-supervised target for the 3D blood vessels by computing the heat kernel signature, a quantity obtained via Laplacian eigenvectors, which captures the very essence of the shapes.
%(MedShapeNet-Blood-Vessel).
We show how geometric representations learned from this datasets can boost segmentation of coronary arteries into regions of low, mid and high (time-averaged) wall shear stress even when trained on limited data.
%This study has implications for foundation models since it isolates the mechanisms of feature extractions and adaption which is a prerequisite to their success.
% 150 to 250 words

%\keywords{First keyword  \and Second keyword \and Another keyword.}
\end{abstract}
%
% Delineate
% - wikipedia.org/wiki/Transfer_learning
% - wikipedia.org/wiki/Generative_pre-trained_transformer
% - wikipedia.org/wiki/Fine-tuning_(deep_learning)
% - wikipedia.org/wiki/Foundation_model
% - wikipedia.org/wiki/Feature_learning
%
% - wikipedia.org/wiki/Self-supervised_learning
% - wikipedia.org/wiki/Unsupervised_learning
%
% Motivate
% - abundance of small specific datasets in medical science
% - e-mail with how researchers want to use my work
%
% - isolate feature extraction by foundation models
% - prerequisite (without emergent properties)
%
% Ideas
% - large-scale epidemiological studies (compare SPARTA)
% - pre-train on subsets of MedShapeNet-Blood-Vessel
% - investigate degradation against only coronary arteries
% - ordinal accuracy metric
% - figure zero
% - add training time per epoch
% - report error and accuracy for LAD, LCX and RCA
% - SO(3) augmentation or equivariance in pre-training
% - stratify artery type balance across folds
%
\section{Introduction}
Cardiovascular disease is an umbrella term for pathological conditions affecting the heart and blood vessels.
Among those, coronary artery disease is the most lethal, e.g.,
% https://www.who.int/news-room/fact-sheets/detail/the-top-10-causes-of-death
causing 9.21 million deaths worldwide in 2021~\cite{Martin2024Heart}.
% Table 21-3. Global Mortality and Prevalence of IHD by Sex, 2021 (p. e797, ch. 21)
Atherosclerosis refers to the accumulation of lipids and inflammatory cells leading to lesions within the artery wall (intima) and is the main pathology of coronary artery disease.
It results in narrowing and hardening of the elastic artery wall which reduces flow of oxygenated blood to the heart muscle with possibly fatal consequences.
Plaque initiation and progression can be linked to (endothelial) shear stress which drives the culpable physiological processes in interaction with the endothelial cells~\cite{Wentzel2022Sex}.
As clinical proxy for this effect, biomarkers such as wall shear stress (WSS) can be used which locally correlates with the atherosclerotic process~\cite{Hoogendorn2019Multidirectional,Hartman2021Definition,DeNisco2024Predicting} and can predict myocardial infarction~\cite{Candreva2022Risk}.
WSS is
% defined as the tangential component of the derivative of blood velocity in direction perpendicular to the artery wall
a first-order differential quantity dependent on the blood flow and can be quantified via computational fluid dynamics (CFD)~\cite{Candreva2022Current,Nannini2024Automated} based on 3D models extracted from subject-specific medical imaging.
However, numerical fluid simulation can be time-consuming.
As faster alternatives, (neural) surrogates have been proposed, that simplify the physical modelling~\cite{Pfaller2022Automated,Pegolotti2024Learning} or learn a geometry-biomarker map from data~\cite{Suk2024Mesh,MoralesFerez2020Deep,Maul2023Transient,Li2021Prediction} rather than solving the governing differential equations.
These methods sacrifice accuracy for inference speed and could be used in iterative or time-critical applications such as \revision{}{screening,} treatment planning or emergency medicine.

While the above neural surrogates demonstrate great potential when trained on sufficiently large datasets \revision{}{(e.g., in the order of thousands of samples)} that are specific to an anatomical region (e.g., coronary arteries, aorta, etc.) with homogeneously generated ground truth data (e.g., via CFD), their generalisation across setups is not yet sufficiently studied.\footnote{The issue of heterogeneous ground truth data is amplified by the fact that CFD is highly sensitive to spatial discretisation and boundary conditions; thus dependent on the person or lab conducting the study~\cite{ValenSendstad2018Real}.}
In biomedical research and engineering, we commonly see small, specific datasets \revision{}{(e.g., in the order of tens or hundreds of samples~\cite{Candreva2022Risk,DeNisco2024Predicting,Nannini2024Automated})} which -- on their own -- are not enough to train powerful neural surrogates but combining which is problematic for privacy reasons.
Furthermore, it is difficult for individual hospitals and research institutes to amass large-enough datasets to train their own models.
In order to nevertheless make use of individual datasets, we can conceive of two possible approaches: (1) federated, task-specific learning \revision{and}{with or without} merging of the trained models in a privacy-preserving manner~\cite{Kaissis2021End} and (2) foundation models \revision{}{(trained on a single large dataset) that can be fine-tuned for each new case}.
Conceptually, foundation models involve large-scale, self-supervised pre-training and produce general-purpose representations to enable the zero- or few-shot behaviour we see in vision or (large) language models.
In both cases, more research is needed about how we can leverage learned representations and adapt models to different contexts.
Cardiovascular simulation data is difficult to standardise across clinics which makes this application particularly challenging.

In this study, we consider the scenario in which an entity has access to a small cohort of patients and would like to leverage self-supervised pre-training for a specific task.
% prerequisite of foundation models
This \revision{is to isolate}{setup isolates} the mechanism of feature \revision{}{learning and} extraction by a foundation model.
We have access to 49 coronary artery models of patients with acute coronary syndrome in which CFD was performed and WSS computed~\cite{DeNisco2021Comparison}.
We pre-train a cross-attention transformer
% (with fixed number of randomly sampled query positions)
on 8449 (3D) point clouds of blood vessel models from MedShapeNet~\cite{Li2025MedShapeNet} with heat kernel signature~\cite{Sun2009Concise} as self-supervised training objective.
The heat kernel signature captures the essence of a shape by means of its manifold structure which makes it promising for geometric representation learning.
We freeze and merge the pre-trained model within a \textbf{G}eometric \textbf{Re}presentation \textbf{A}daption \textbf{T}ransformer (GReAT) that classifies regions of low, mid and high (time-averaged) WSS in the coronary artery models.
We find that adapting the geometric representation to this specific task can boost accuracy with little to no overhead in training or inference.
We open source our implementation as well as publish the subset MedShapeNet-Blood-Vessel (including heat kernel signature).\footnote{\revision{\texttt{github.com/available-upon-publication}}{\href{https://github.com/sukjulian/great}{github.com/sukjulian/great}}}
Our contributions are (1) GReAT, i.e., laying out a framework for leveraging learned geometric representations on unrelated tasks, (2) curating a large dataset of blood vessels from MedShapeNet and (3) proposing (and computing) heat kernel signature as self-supervised training objective.

\section{Related works}

In the following we provide an overview over select related works in the context of task-specific leveraging of previously learned representations.

\subsubsection{Self- and unsupervised pre-training for 3D point clouds}
Most methods fall into one of two categories: contrastive or generative.
Contrastive learning is done by identifying augmented instances of the same point cloud at training time while discriminating from different point clouds.\footnote{Here, augmentations mean random transformations applied to the point cloud\revision{}{, such as rotations and scaling}.}
Approaches use the contrastive mechanism for unsupervised pre-training and then fine-tune the models in a supervised way.
This has mainly been explored for point clouds of 3D scenes~\cite{Xie2020PointContrast,Zhang2021Self,Huang2021Spatio}.
Notably though, Liu et al.~\cite{Liu2023Hierarchical} proposed self-supervised contrastive pre-training for 3D meshes of human teeth.
Among other augmentations, they apply rigid transformations from $\mathrm{SE}(3)$.
In contrast to contrastive learning, generative approaches typically disturb the point cloud and set its reconstruction as pre-training objective.
Yu et al.~\cite{Yu2022Point} propose masking out local regions followed by point cloud completion (from fixed latent space) via generative autoencoders.
Sauder er al.~\cite{Sauder2019Self} relax the architecture requirement and propose spatially rearranging and then reconstructing voxels of points using any (permutation-equivariant) point cloud model.
Apart from 3D point clouds, self-supervised pre-training has been applied to 3D medical images~\cite{Tang2022Self}.
In contrast to these works, we explore 3D point clouds of human blood vessels in this study; we propose a simple regression objective that leverages the \textit{intrinsic} structure of the manifolds underlying these point clouds; we evaluate our framework on the assessment of clinically relevant biomarkers.

\subsubsection{Self-supervised pre-training on spectral features}
In the context of graph representation learning, Cantürk et al.~\cite{Cantürk2024Graph} proposed a self-supervised regression objective including graph Laplacian eigenvectors and derived heat kernel signature.
The learned representation can be used as general-purpose feature for downstream tasks.
Analogously to graphs, Laplacians can be defined on meshes and point clouds with similar implications, which we propose in this study.

\subsubsection{Foundation models in medicine}
Models trained on extremely large datasets in a self- or unsupervised manner have caught the interest of the medical community.
Large language models are the primary modality of state-of-the-art foundation models but are currently lacking precision for clinical decision-making~\cite{Hager2024Evaluation}.
In contrast, image segmentation models like MedSAM~\cite{Ma2024Segment} are proving useful and are gradually being integrated in medical image segmentation workflows.
In this study, we aim to contribute a suitable framework to the available tools for geometric foundation models in cardiovascular medicine.
We also provide empirical evidence of the efficiency \revision{}{(}of their underlying mechanism\revision{}{)} of transfer learning.

\section{Datasets}
In the following we present details about the large, generic and small, specific datasets we use in this study.

\begin{figure}[t]
	\centering
	\includegraphics{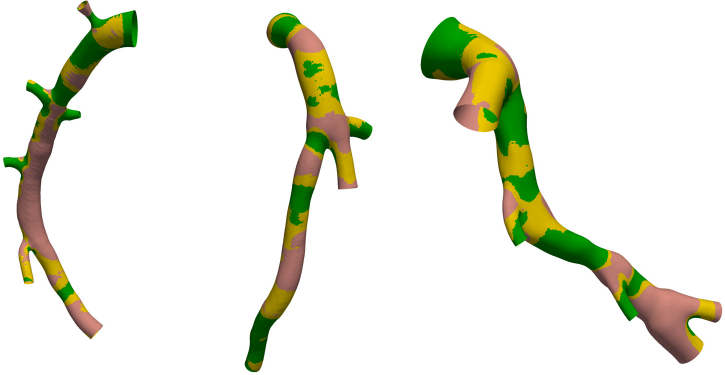} % 347.12354 pt
	\caption{\textbf{Coronary arteries} from the IMPACT dataset. Right coronary artery (left), left anterior descending (middle) and left circumflex (right) with regions of low (pink), mid (yellow) and high (green) WSS.}
	\label{fig:impact}
\end{figure}

\subsection{Coronary artery wall shear stress}
The IMPACT study~\cite{DeNisco2021Comparison} comprised 49 coronary arteries in patients with acute coronary syndrome.
The coronary arteries were scanned (via intravascular ultrasound and computed tomography angiography) and segmented, resulting in 3D (triangular) surface meshes with around 70k points.
The data contains variation in anatomical location: 18 left anterior descending (LAD), 13 left circumflex (LCX) and 18 right coronary arteries (RCA) (see Fig.~\ref{fig:impact}).
Previously obtained WSS extracted from CFD simulation acts as the estimation objective.
For each patient, personalised boundary conditions are included in the form of an inflow waveform.
Based on the arguments by Hartman et al.~\cite{Hartman2021Definition}, we classify the WSS into low, mid and high values by applying a balanced tertile threshold \textit{per artery}.\footnote{Regions of low WSS correlate with high risk of plaque progression. This effect is most pronounced when a per-artery (opposed to a static or study-wide) threshold is used.}
To this end, we normalise the time-dependent WSS vectors and subsequently average over time to obtain time-averaged WSS.
We thus turn WSS estimation into a three-class segmentation problem which greatly simplifies the task (and interpretability of evaluation metrics)\revision{ at no loss of clinical relevance}{}.
% we excluded three patients due to misterious reasons

\begin{figure}[t]
	\centering
	\includegraphics{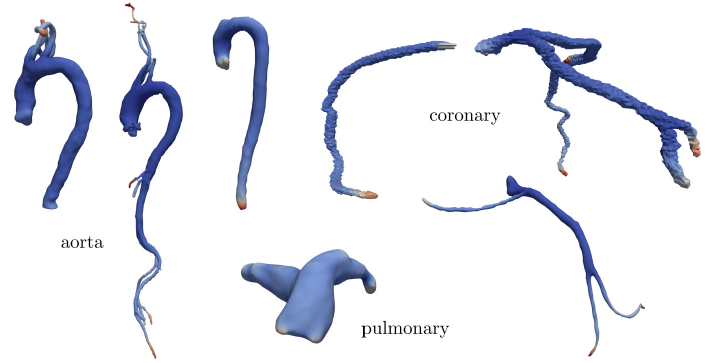} % 347.12354 pt
	\caption{\textbf{MedShapeNet-Blood-Vessel.} Shown are examples of three classes with heat kernel signature. While blood vessels in this dataset have different levels of detail (see aorta) and mesh quality (see coronary), heat kernel signature is robust across examples.}
	\label{fig:medshapenet}
\end{figure}

\subsection{MedShapeNet-Blood-Vessel}
MedShapeNet~\cite{Li2025MedShapeNet} is a large dataset of anatomical shapes, represented as 3D surface meshes.
For this study, we manually extract all shapes related to blood vessels and name the resulting subset MedShapeNet-Blood-Vessel.
These include the coronary arteries, thoracic aorta, celiac trunk, abdominal aorta, iliac arteries, iliac veins, portal vein, splenic vein, inferior vena cava and pulmonary artery.
The shapes are represented by heterogeneous sets of point coordinates (average 13k, maximum 32k)
% minimum 130
and triangular faces.
We standardise the point coordinates by first subtracting the center of mass and afterwards dividing by the standard deviation across all points' distances to the origin.\footnote{This procedure facilitates comparing the point clouds to any new point cloud at the cost of erasing explicit scale information.}
For each of these shapes we compute a \textit{point cloud} Laplacian (i.e., disregarding face information) with an open-source Python package~\cite{Sharp2020Laplacian}, adding an extra layer of stability in case of noisy meshes.
Subsequently, we \revision{solve for 128 pairs of eigenvalue and eigenvector}{compute eigenvectors corresponding to the 128 lowest eigenvalues of the (positive definite) Laplacian} and compute the heat kernel signature for 16 time points between \num{1e-2} and \num{1e0} (see \ref{sec:hks}).
Solving the eigenvalue problem failed in six cases which we exclude, leaving 8449 shapes in total.
We standardise the heat kernel signature by subtracting the mean and dividing by standard deviation across (space) points for each time point separately.

\section{Methods}
\revision{In the following we will discuss some}{The following section provides} necessary background and the main methodological contribution of this study.

\begin{figure}[t]
	\centering
	\includegraphics{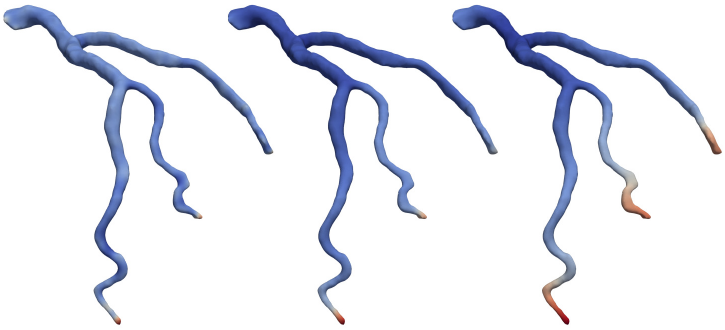} % 347.12354 pt
	\caption{\textbf{Heat kernel signature} for coronary arteries from MedShapeNet-Blood-Vessel at three points in time. The heat kernel signature provides multiscale structural encoding where time points $t$ correspond to different scales. This correlates with Gaussian curvature for small $t$~\cite{Sun2009Concise}.}
	\label{fig:hks}
\end{figure}

\subsection{Heat kernel signature}\label{sec:hks}
Consider an approximation $L \in \mathbb{R}^{n \times n}$ to the Laplace-Beltrami operator on a 2D manifold discretised by $n$ points in 3D.\footnote{The Laplace-Beltrami operator is used within the heat equation to simulate diffusion of a heat source on a manifold.}
Let $L$ be symmetric and positive definite and denote its eigendecomposition by $L = W \Lambda W^\mathsf{T}$ were columns of
\[
W = \begin{pmatrix} | & & | \\ w^1 & \cdots & w^n \\ | & & | \end{pmatrix}
\]
are $n$-dimensional (unit) eigenvectors and $\Lambda = \diag(\lambda_1, \dots, \lambda_n)$ are eigenvalues.
Discrete heat kernel signature $\mathrm{HKS} \in \mathbb{R}^n$ at time $t > 0$ is defined as
\[
\mathrm{HKS}_i = \sum_{j=0}^{n} e^{-\lambda_i t} (w^j_i)^2.
\]
This vector can be mapped to the points of a discrete manifold (see Fig.~\ref{fig:hks}) and is a powerful intrinsic structural representation.
For more details, we refer to Sun et al.~\cite{Sun2009Concise}.
We propose heat kernel signature as supervision target for self-supervised pre-training, since it can be computed purely from point coordinates of a discretised manifold.
\revision{}{In particular, every manifold surface mesh in 3D admits construction of a smooth atlas (of charts) which in turn enables definition of a Riemannian metric and Laplace-Beltrami operator.}

\subsection{Model architecture}
In this study, we use an ablation to the LaB-GATr~\cite{Suk2024LaB,Suk2024Geometric} model by replacing its components operating under geometric algebra with \textbf{Va}nilla layers (LaB-VaTr) operating under linear algebra.
Lab-VaTr is closely related to UPT~\cite{Alkin2024Universal} and Perceiver IO~\cite{Jaegle2022Perceiver}.
The model is composed of a cross-attention encoder which feeds into multiple self-attention blocks, followed by a decoder based on learned interpolation.
The encoder pools a point cloud of input features to a coarse subset of the points with controllable size.
Afterwards, the self-attention module operates on this latent representation.
The decoder recovers input resolution by -- for each input point -- interpolating between the three closest coarse points (Euclidean metric).
The decoder is equivalent to PointNet++~\cite{Qi2017PointNet++} and has been observed to improve performance compared to cross-attention~\cite{Suk2024Geometric}.
% consider providing the formula
Coarse representations are randomly sampled from the point clouds.\footnote{Note that this means we have to compute nearest neighbours for coarse points at each forward pass. Modern implementations on the GPU compute all pairwise distances between coarse and fine points, resulting in similar computational complexity $\mathcal{O}(n_\text{coarse} n_\text{fine})$ as cross-attention.}
The proposed model is highly scalable due to its controllable latent representation.
Computational bottlenecks are encoder and decoder both of which scale sub-quadratically in the number of input points.
Nevertheless, for the purpose of large-scale pre-training, we skip the interpolation layer altogether and learn the supervision target mapped to the coarse points only.

\subsection{Fine-tuning}
We leverage self-supervised pre-training by fusing the learned representation in a \textbf{G}eometric \textbf{Re}presentation \textbf{A}daption \textbf{T}ransformer (GReAT).
To this end, we set up two instances of LaB-VaTr
\[
f_i(\cdot) = (\delta_i \circ \beta_i \circ \epsilon_i)(\cdot), \hspace{12pt} i \in \{1, 2\},
\]
both composed of an encoder $\epsilon_i$ that maps from $(n_\text{fine} \times c_{\text{in}, i})$ to $(n_\text{coarse} \times c_\beta)$ tensors, a sequence of intermediate blocks (coalesced in) $\beta_i$ and a decoder $\delta_i$ that maps from $(n_\text{coarse} \times c_\delta)$ to $(n_\text{fine} \times c_{\text{out}, i})$ tensors.
We pre-train $f_1$ and freeze all its trainable parameters afterwards.
Denote by $\Vert$ concatenation of tensors along the second (or channel) dimension as an operator that can be applied to functions.
We compose
\[
\mathrm{GReAT} = \delta_2 \circ \beta_2 \circ (\epsilon_2 \Vert (\beta_1 \circ \epsilon_1)),
\]
i.e., concatenate the output of $\beta_1$ as learned geometric representation with $\epsilon_2$.
Both $\epsilon_1$ and $\epsilon_2$ receive the same point cloud as input in our architecture which is why we standardise the point coordinates by first subtracting the center of mass and afterwards dividing by the standard deviation across all points' distances to the origin.

\section{Experiments}
We set up all LaB-VaTr (sub-)models with eight self-attention blocks and 128 hidden channels, so that they have just over 1M parameters.
As pooling target for the encoder we choose random subsets of size $n_\text{coarse} = 1000$ in each forward pass.
We use Adam with learning rate \num{3e-4} for training.
\revision{}{We train all models on an Nvidia A40 GPU.}

\subsection{Pre-training}
We train a LaB-VaTr model (dropping the decoder module) on MedShapeNet-Blood-Vessel with heat kernel signature as supervision target, $\mathrm{L}^1$ loss and batch size 256.
Inputs are point coordinates only.
We observe convergence on a held-out validation split of 848 anatomies after 400 epochs \revision{}{(ca. 15 h 32 min)}.
% (13600 gradient updates)
\revision{}{Inference takes ca. 0.0147 s per shape in MedShapeNet-Blood-Vessel and ca. 0.2203 s per shape in IMPACT.}
In Table~\ref{tab:hks} we report accuracy metrics on the (balanced) test split of 848 anatomies as well as the full IMPACT dataset.\footnote{We publish our splits alongside MedShapeNet-Blood-Vessel.}
Even though MedShapeNet-Blood-Vessel contains coronary artery shapes, we see slight degradation of accuracy (0.21 $\to$ 0.38 MAE) in prediction of heat kernel signature for IMPACT.

\begin{table}[t]
	\centering
	\caption{\textbf{Mean absolute error} (MAE) for heat kernel signature prediction by LaB-VaTr trained on MedShapeNet-Blood-Vessel. We report mean $\pm$ standard deviation across subjects.}
	\begin{tabular}{lc}
		\toprule
		Dataset & MAE $\downarrow$ \\
		\midrule
		MedShapeNet-Blood-Vessel$^\dagger$ & 0.21 $\pm$ 0.16 \\
		IMPACT$^\ddagger$ & 0.38 $\pm$ 0.1\hphantom{0} \\
		\bottomrule
		\multicolumn{2}{l}{$^\dagger$ test split $^\ddagger$ full dataset}
	\end{tabular}
	\label{tab:hks}
\end{table}

\subsection{Fine-tuning}
\begin{figure}[t]
	\centering
	\includegraphics{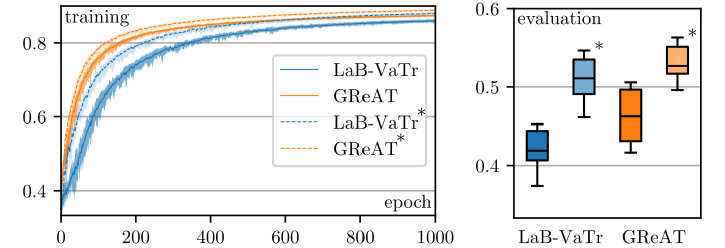} % 347.12354 pt
	\caption{\textbf{Classification accuracy $\uparrow$} (number of correctly classified points divided by total) for IMPACT. We show metrics across eight-fold cross-validation. Models annotated by $^*$ use local artery radius as additional input.}
	\label{fig:cross_validation}
\end{figure}

We train GReAT (using the learned geometric representation) on IMPACT for three-class (low, mid and high) WSS-region segmentation under cross-entropy loss.
Inputs are point coordinates, surface normal, statistics about the waveform (mean and standard deviation), type of the artery (LAD, LCX or RCA) and geodesic distance of each point to the artery inlet and outlet.\footnote{Geodesic distance to a source can be computed on meshes via the heat method~\cite{Crane2017Heat}.}
Due to the small size of the dataset we perform eight-fold cross-validation where we evaluate on six arteries after training on all others.\footnote{We make sure that arteries of the same patient are not scattered across training and evaluation split.}
Batch size is eight and we train for 1000 epochs \revision{}{(ca. 2 h 47 min)} with exponential learning rate decay of 0.9977.
In order to quantify the effect of pre-training, we also train a baseline LaB-VaTr with these same settings.
In Fig.~\ref{fig:cross_validation} we show training and test accuracy which we defined as the number of correctly classified points divided by the total number of points.
\revision{}{Note that the classes are perfectly balanced by construction.}
We observe that GReAT converges faster and to a higher accuracy than the baseline.
Furthermore, GReAT is significantly ($p = 0.016$) more accurate on the test split.\footnote{We test significant deviation between population means via one-way ANOVA test.}
For both models, accuracy during training is higher than for evaluation, pointing towards overfitting (note that 0.33 corresponds to random guessing).

\subsubsection{Local artery radius as input}
\begin{figure}[t]
	\centering
	\includegraphics{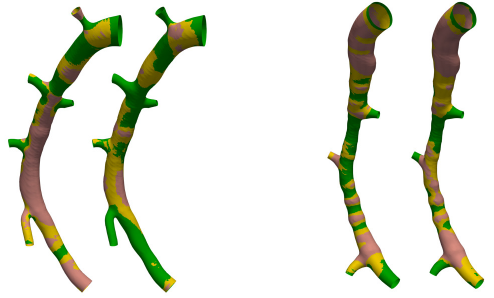} % 347.12354 pt
	\caption{\textbf{Segmentation} into regions of low (pink), mid (yellow) and high (green) WSS. Labels (left, respectively) and model predictions (right, respectively) for subjects corresponding to lowest (left, 0.5) and highest (right, 0.56) evaluation accuracy. Both cases are right coronary arteries.}
	\label{fig:wss}
\end{figure}

In order to study task-specific limits of self-supervised pre-training we add a morphological input feature that highly correlates with WSS: local artery radius.
We define it as normed relative position to the centreline which we compute via skeletonisation.
Fig.~\ref{fig:cross_validation} shows training and test accuracy.
Both GReAT and the baseline benefit from the additional feature, but the effect is more pronounced for the latter.
Training accuracy for the models is still clearly separated into two modes, suggesting efficacy of leveraging the learned geometric representation.
However, there is no longer significant ($p = 0.166$) difference in evaluation accuracy.
In Fig.~\ref{fig:wss} we show examples of WSS-region segmentation by GReAT.

\subsubsection{Geometric input features ablation}
We investigate the quality of the geometric representation learned during pre-training by ablating all geometric input features, i.e., point coordinates and surface normal, for both models.
GReAT still builds a geometric representation from the point coordinates, but there is no trainable pathway from those to the model output.
This means GReAT has to rely solely on the learned representation for geometric context.
We use local artery radius as additional input to create a geometry-free but morphologically-informed baseline.
We present training and evaluation accuracy in Table~\ref{tab:ablations}.
Without geometric context, LaB-VaTr converges to far lower training accuracy (0.87 $\to$ 0.76) compared to GReAT.
GReAT has significantly ($p = 0.03$) higher evaluation accuracy than LaB-VaTr, both suggesting the learned representation is a useful prior.
However, addition of point coordinates and surface normal ("Local artery radius as input") significantly ($p = 0.001$) increases evaluation accuracy, which shows that these features are nevertheless valuable.

\begin{table}
	\centering
	\caption{\textbf{Ablation studies} for IMPACT. We report the mean for cross-validation folds. Where evaluation accuracy is \underline{underlined} it is significantly ($p < 0.05$) different between both models. We denote best evaluation accuracy in \textbf{bold}. Note that "Geometric input features ablation" uses local artery radius as additional input.}
	\begin{tabular}{llcc}
		\toprule
		\multirow{2}{*}{Modification} & \multirow{2}{*}{Model} & \multicolumn{2}{c}{Accuracy $\uparrow$} \\
		\cmidrule(l){3-4}
		& & Training & Evaluation \\
		\midrule
		\multirow{2}{70pt}{--} & LaB-VaTr & 0.86 & 0.42 \\
		& GReAT & 0.87 & \underline{0.46} \\
		\midrule
		\multirow{2}{70pt}{Local artery radius as input} & LaB-VaTr & 0.88 & \textbf{0.51} \\
		& GReAT & 0.89 & \textbf{0.53} \\
		\midrule
		\multirow{2}{70pt}{Geometric input features ablation} & LaB-VaTr & 0.76 & 0.46 \\
		& GReAT & 0.87 & \underline{0.49} \\
		\midrule
		\multirow{2}{70pt}{Computed heat kernel signature} & LaB-VaTr & 0.90 & 0.47 \\
		& GReAT & 0.90 & 0.49 \\
		\bottomrule
	\end{tabular}
	\label{tab:ablations}
\end{table}

\subsubsection{Computed heat kernel signature}
Finally, we check how pre-training with heat kernel signature as self-supervised target compares to explicitly computing it per subject.
To this end, we choose the same inputs as the base experiment ("--") and additionally pass heat kernel signature, computed as described in \ref{sec:hks}.
While this leads to significant ($p = 0.005$) increase in evaluation accuracy for LaB-VaTr, it causes no significant ($p = 0.053$) change for GReAT.
When comparing base experiment GReAT to LaB-VaTr using heat kernel signature as additional input, we find no significant ($p = 0.747$) difference in evaluation accuracy.
From these tests we conclude that the learned geometric representation is powerful enough to make explicitly computed heat kernel signature obsolete for this experiment.

\section{Discussion}
In this study, we show how self-supervised pre-training on purely geometric data can be leveraged for a specific task involving cardiovascular biomarkers.
We attribute this to the fact that geometry-based neural networks learn mappings from 3D geometries to the task-specific labels and necessarily must create a geometric representation of the input.
In particular, we study the assessment of WSS in a small cohort of patients.
We show that learning said representation can be outsourced to self-supervised pre-training without the need for task-specific labels.
This enables us to leverage MedShapeNet-Blood-Vessel, a (new) large dataset of anatomical shapes, while circumventing their (expensive) labelling, i.e., CFD simulation.

Efficacy of large-scale pre-training on a self-supervised task is a precursor to foundation models.
In the context of blood vessel shapes, we show that our scalable transformer model can learn useful representations from geometric data and transfer it to a downstream task.
We envision our framework being scaled up in terms of \revision{dataset size}{available training data} and model parameters and ultimately being developed into a \revision{geometric}{general-purpose} foundation model for cardiovascular \revision{biomarkers}{representation learning}.
\revision{}{The foundation model could then be fine-tuned by hospitals and research institutes for in-house use, circumventing concerns regarding patient privacy.
This could provide novel decision markers (such as low time-averaged WSS) for clinical prognosis and treatment planning (e.g., atherosclerosis prevention) considering which could ultimately benefit patient well-being.}

In this study, we use heat kernel signature as self-supervision target because it captures the essence of a shape across scales and anatomical variation well.
Cantürk et al.~\cite{Cantürk2024Graph} also proposed heat kernel signature, among others, for this purpose but their study was focussed on graphs rather than 2D manifolds.
We propose a scalable, transformer-based neural network architecture which we use to learn from and across large-scale anatomical point clouds.
Our approach of fusing the learned geometric representation with the task-specific one is similar to a "skip connection", allowing the model to disregard the former if it is optimal.
We show that our framework -- GReAT -- enables zero-shot generalisation by improving segmentation accuracy compared to the baseline.
Furthermore, we show in an ablations study that the learned geometric representation alone suffices to make GReAT converge during training even without a trainable pathway from labels to point coordinates.
This supports the claim that the model has learned to extract a useful geometric representation from self-supervised pre-training.

In future work, we aim to compare our self-supervised approach to baselines built on training with contrastive loss as well as masked autoencoders.
Furthermore, it would be interesting to also study supervised pre-training and transfer learning on task-specific WSS data.

\section{Limitations}
In the following we discuss some limitations of this study.
After pre-training on MedShapeNet-Blood-Vessel, we observe degradation of test error when estimating heat kernel signature on IMPACT, even though the former contains coronary arteries.
The reasons could be that the model overfits to 3D representation artefacts such as the meshing algorithm or that the model is not optimal for surface structures.
\revision{}{In order to mitigate this domain sensitivity and enable clean domain transfer, one could homogenise source and target domain via re-meshing and canonical alignment or pre-train under data augmentation.}
From our experiment with local artery radius as input, we find that the impact of self-supervised pre-training is limited in presence of a highly expressive feature.
This is likely due to the limited size of the task-specific dataset: the model is not fully capable of explaining WSS regions by surface geometry alone and is overfitting to the local artery radius.

\section{Conclusion}
We find that self-supervised pre-training on shape data can be leveraged for a downstream task involving shape-informed, hemodynamic biomarkers.
We perform several ablations to understand the efficacy and failure modes of this mechanism.
In conclusion, we see
% great
potential for self-supervised pre-training on large dataset in the context of 3D geometries and cardiovascular biomarkers.

\begin{credits}
\subsubsection{\ackname}
Jelmer M. Wolterink was supported by the NWO domain Applied and Engineering Sciences VENI grant (18192).

\subsubsection{\discintname}
The authors have no competing interests to declare that are
relevant to the content of this article.
\end{credits}


\bibliographystyle{splncs04}
\bibliography{mybibliography}

\end{document}